\documentclass[1p]{elsarticle}

\usepackage{amssymb}
\usepackage{amsmath}
\usepackage{graphicx}
\usepackage{lineno}
\usepackage{siunitx}
\usepackage{tipa}  
\usepackage{booktabs}  

\usepackage{hyperref}  

\title{Decoding Phone Pairs from MEG Signals Across Speech Modalities}

\author[inst1]{Xabier de Zuazo\corref{cor1}}
\ead{xabier.dezuazo@ehu.eus}

\author[inst1]{Eva Navas}
\ead{eva.navas@ehu.eus}

\author[inst1]{Ibon Saratxaga}
\ead{ibon.saratxaga@ehu.eus}

\author[inst2,inst4,inst5,inst6]{Mathieu Bourguignon}
\ead{mabourgu@ulb.ac.be}

\author[inst2,inst3]{Nicola Molinaro}
\ead{n.molinaro@bcbl.eu}

\affiliation[inst1]{organization={HiTZ Center, University of the Basque Country -- UPV/EHU},
    addressline={Ingeniero Torres Quevedo Plaza 1},
    postcode={48013},
    postcodesep={},
    city={Bilbao},
    country={Spain}}

\affiliation[inst2]{organization={Basque Centre on Cognition, Brain and Language -- BCBL},
    addressline={Paseo Mikeletegi 69},
    postcode={20009},
    postcodesep={},
    city={Donostia - San Sebastián},
    country={Spain}}
\affiliation[inst3]{organization={Ikerbasque, Basque Foundation for Science},
    addressline={Plaza Euskadi 5},
    postcode={48009},
    postcodesep={},
    city={Bilbao},
    country={Spain}}

\affiliation[inst4]{organization={Laboratory of Functional Anatomy, Faculty of Human Motor Sciences, Université libre de Bruxelles (ULB)},
    addressline={Route de Lennik 808, CP 619},
    postcode={1070},
    city={Brussels},
    country={Belgium}}
\affiliation[inst5]{organization={Laboratoire de Neuroanatomie et Neuroimagerie translationnelles (LN2T), ULB Neuroscience Institute, Université libre de Bruxelles (ULB)},
    addressline={Route de Lennik 808},
    postcode={1070},
    city={Brussels},
    country={Belgium}}
\affiliation[inst6]{organization={WEL Research Institute},
    addressline={Avenue Pasteur 6},
    postcode={1300},
    city={Wavre},
    country={Belgium}}

\cortext[cor1]{Corresponding author}

\journal{Computer Speech \& Language}

\begin{document}


\begin{abstract}
    Understanding the neural mechanisms underlying speech production is essential for both advancing cognitive neuroscience theory and developing practical communication technologies. In this study, we investigated magnetoencephalography (MEG) signals to decode phonetic units (phones) from brain activity during speech production and perception (passive listening and voice playback) tasks. Using a dataset comprising 17 participants, we performed pairwise phone classification, extending our analysis to 15 phonetic pairs. Multiple machine learning approaches, including regularized linear models and neural network architectures, were compared to determine their effectiveness in decoding phonetic information. Our results demonstrate significantly higher decoding accuracy during speech production (76.6\%) compared to passive listening and playback modalities (approximately 51\%), emphasizing the richer neural information available during overt speech. Among the models, the Elastic Net classifier consistently outperformed more complex neural networks, highlighting the effectiveness of traditional regularization techniques when applied to limited and high-dimensional MEG datasets. Besides, analysis of specific brain frequency bands revealed that low-frequency oscillations, particularly Delta (\SIrange{0.2}{3}{\hertz}) and Theta (\SIrange{4}{7}{\hertz}), contributed the most substantially to decoding accuracy, suggesting that these bands encode critical speech production-related neural processes. Despite using advanced denoising methods, it remains unclear whether decoding solely reflects neural activity or if residual muscular or movement artifacts also contributed, indicating the need for further methodological refinement. Overall, our findings underline the critical importance of examining overt speech production paradigms, which, despite their complexity, offer opportunities to improve brain-computer interfaces to help individuals with severe speech impairments.
\end{abstract}

\begin{graphicalabstract}
\includegraphics[width=\linewidth]{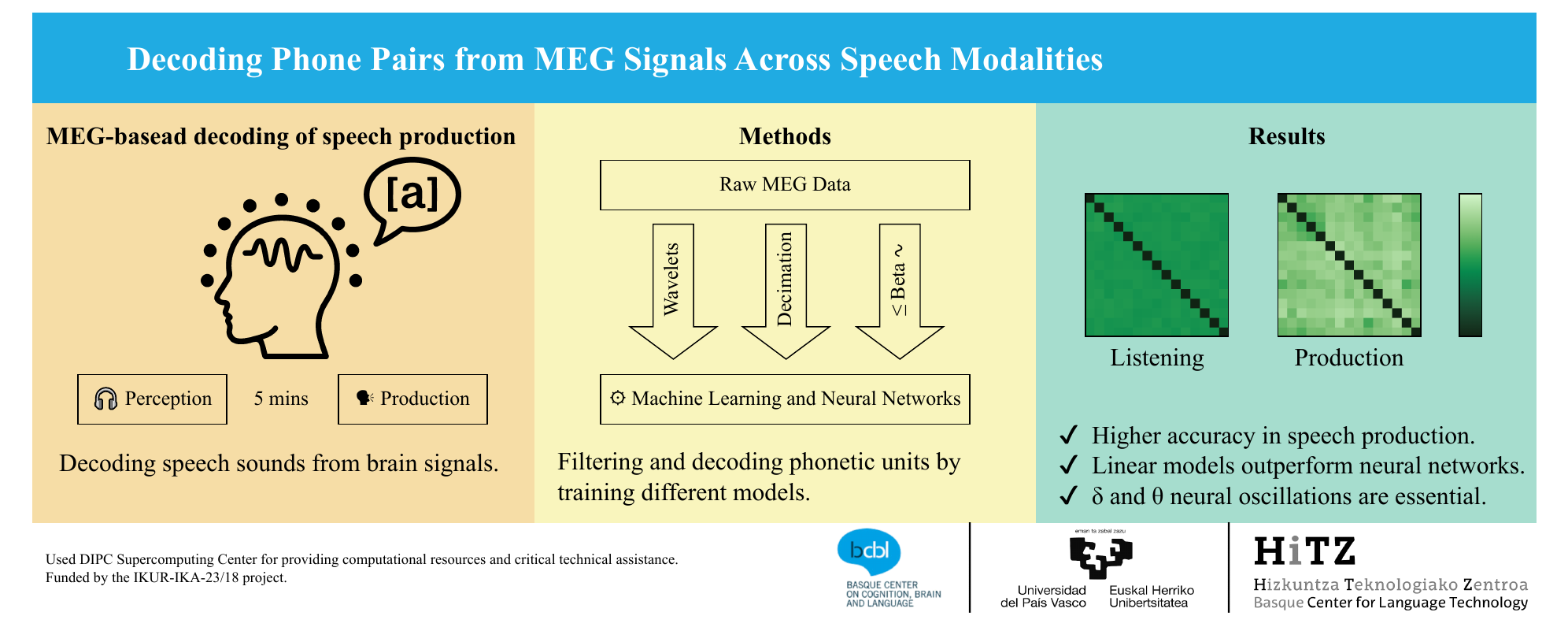}
\end{graphicalabstract}

\begin{highlights}
\item MEG decoding of 15 phonetic units during speech production.
\item Higher decoding accuracy in speech production vs. perception.
\item Linear models outperform neural networks on limited MEG data.
\item Delta and Theta bands are essential for decoding speech production.
\end{highlights}


\maketitle

\section{Introduction}

Brain-computer interfaces (BCIs) have emerged as essential technologies, facilitating direct interactions between neural activity and external devices. BCIs offer unique opportunities to improve the quality of life for individuals with severe communication impairments caused by conditions such as neurodegenerative diseases or paralysis~\cite{Anumanchipalli2019, Moses2019, Silva2024}. Among various non-invasive neuroimaging methods employed for BCI development, functional Magnetic Resonance Imaging (fMRI) is a technique that tracks changes in cerebral blood flow, offering high spatial resolution but limited temporal resolution, thus restricting its application in real-time communication devices~\cite{Logothetis2008}. Electroencephalography (EEG), a non-invasive method widely employed due to its ease of use and high temporal resolution, captures electrical activity generated by neurons at the scalp level but has considerable spatial resolution limitations and signal-to-noise ratio challenges~\cite{Burle2015, Sato2024}. In contrast, magnetoencephalography (MEG), another non-invasive approach, measures magnetic fields generated by neuronal activity. MEG combines excellent temporal resolution with spatial accuracy superior to EEG, though it demands specialized infrastructure and has higher operational costs~\cite{Cohen1983, Destoky2019}.

Speech production can be compromised by neurological conditions, such as Alzheimer's disease or Primary Progressive Aphasia, which progressively deteriorate the complex neural networks governing spoken communication. Although addressing speech production impairments is very important, neuroscientific research has focused predominantly on passive speech perception paradigms, such as reading or listening tasks~\cite{Schoffelen2019, Nastase2021, Gwilliams2023}, mainly due to the presence of motor artifacts during overt speech tasks. Consequently, studies on speech production have often relied on unnatural experimental paradigms, which raises concerns regarding their ecological validity. As a result, a significant gap persists in neural data acquisition and analytical methodologies for overt speech production. Public datasets containing audio synchronized with neural signals during speech production tasks are extremely limited. Yet, such datasets are essential for developing practical speech-based BCIs that aim to decode intended speech directly from neural signals rather than interpreting passive perception of speech~\cite{Philip2022, Philip2023, Levy2025}. Bridging this gap by intensifying research efforts in speech production can substantially improve technological support for individuals experiencing profound communication difficulties.

Recent advances have substantially developed our understanding of how speech-related neural processes can be decoded into usable signals for communication. Studies employing invasive neural recordings have demonstrated the possibility of translating neural activity into comprehensible speech or text, improving BCI power for speech restoration~\cite{Herff2015, Anumanchipalli2019, Livezey2019, Moses2019, Silva2024}. In parallel, non-invasive techniques, particularly MEG, have been increasingly explored due to their ethical and safety advantages. Recent pioneering work by Kwon et al.~\cite{Kwon2024} has shown the feasibility of synthesizing intelligible speech directly from MEG signals, highlighting an essential breakthrough for non-invasive BCIs. Furthermore, extensive research efforts have explored the use of EEG for open vocabulary speech decoding, emphasizing the importance of dataset size and representation learning in achieving reliable decoding performance~\cite{Sato2024}. Additional studies have demonstrated decoding capacity at word and phonetic levels using non-invasive recordings across large populations~\cite{dAscoli2024}, emphasizing the importance of considering factors like data quantity, modality (e.g., MEG versus EEG), and experimental paradigms (reading versus listening). Notably, previous studies by Dash et al. have demonstrated effective decoding of speech production at the word level from MEG signals using deep learning classification methods, achieving successful discrimination across five different classes~\cite{Dash2018_1, Dash2018_2, Dash2019}. Such studies underscore the feasibility of using MEG-based decoding approaches customized explicitly to overt speech production, an area still relatively underexplored compared to speech perception.

Building upon our previous preliminary work~\cite{DeZuazo24}, the current study extends the scope by increasing the set of analyzed phonetic pairs from 10 to 15, providing a more complete phonetic decoding analysis using MEG data during speech production and perception tasks. We cover a broader family of models, including traditional machine learning and more recent neural networks. Moreover, we include a systematic analysis of classical brain frequency bands to assess the frequency ranges that carry phonetic information across tasks. This was performed to guide and justify the use of low-pass filtering during MEG preprocessing for model training. Prior work with the same dataset~\cite{Bourguignon2020} emphasized the role of low-frequency neural oscillations, particularly Delta and Theta bands, in speech processing, and our frequency band analysis extends this to explicitly examine their contribution to decoding performance. In addition, we performed an ablation study to evaluate the individual contribution of each preprocessing component, helping to systematically justify the final decoding pipeline used across all model comparisons. Through this analysis, we aim to better understand how neural activity patterns vary between overt speech production and passive listening, revealing commonalities and distinctions that can guide future speech-BCI research. This extended work studies the possibilities of integrating non-invasive neuroimaging techniques with detailed phonetic-level analysis to improve communication aids for individuals with impaired speech. Besides, we share our fully reproducible pipeline source code\footnote{\url{https://github.com/hitz-zentroa/meg-phone-decoding}}
that other researchers can adapt for further research.

\section{Materials and Methods}

\subsection{Dataset}

The current study employs the dataset previously described by Bourguignon et al. (2020)~\cite{Bourguignon2020}. Initially, 18 healthy adult participants, all native Spanish speakers, participated in the MEG recordings. One participant was later excluded from analyses due to excessive artifacts in the MEG signals. Therefore, the present work considers data from 17 participants, comprising nine females and eight males, aged between 20 and 32 years (mean: 23.9 years). Most participants (16 out of 17) were right-handed, as determined by the Edinburgh Handedness Inventory (mean laterality quotient: 70.6 \%, range: 40--100\%;~\cite{Oldfield1971}), while handedness data was unavailable for one participant. Among the participants, 13 had completed a university degree, one was enrolled as a master's student, and three had completed secondary or high school education. All participants had no reported history of neurological, psychiatric, or language-related disorders and possessed normal hearing and speech abilities. Ethical approval for the study was granted by the Ethics Committee of the Basque Center on Cognition, Brain and Language (BCBL). All participants provided informed written consent before participation.

\subsection{Experimental Procedures and MEG Recordings}

MEG signals were collected at the BCBL facilities using a 306-channel Elekta Neuromag scanner (Vectorview \& Maxshield™; MEGIN Elekta Oy, Helsinki, Finland) housed in a magnetically shielded room (MSR). The MEG system comprises 102 sensor units, each containing one magnetometer and two orthogonal planar gradiometers, totaling 102 magnetometers and 204 gradiometers. This sensor configuration provides complementary sensitivity patterns, which are advantageous for detecting both superficial and deeper cortical sources.

Neuromagnetic signals were continuously sampled at \SI{1}{\kilo\hertz}, with an acquisition band-pass range set between \SIrange{0.1}{330}{\hertz}. Five head-position indicator coils were placed on each participant's scalp to continuously track the participant's head position throughout the recording. These coils generated identifiable magnetic fields captured by the MEG sensors, facilitating accurate tracking of head movements. For spatial co-registration purposes, anatomical landmarks (nasion, left and right preauricular points) and at least 150 additional scalp and nose surface points were digitized using a Polhemus Fastrak electromagnetic tracker (Polhemus, Colchester, VT, USA).

Participants performed three speech-related tasks: reading aloud, listening to speech, and listening to a playback of their previously recorded speech. Each task was carried out for approximately \SI{5}{\minute}, employing two distinct narrative texts in Spanish, each containing roughly 1000 words (average speaking pace: 2.53 \textpm{} 0.12 words per second). During speech production, participants read one of the texts aloud, printed on A4 paper. The speech was simultaneously recorded through an optical fiber microphone (sampling rate: \SI{44.1}{\kilo\hertz}) placed approximately \SI{5}{\centi\meter} from the participant's mouth, attached to the edge of the MEG helmet. In the listening condition, participants heard a pre-recorded speech by a native speaker matching their gender. For the playback task, they listened to the audio of their previously recorded speech. Sound stimuli for listening and playback conditions were delivered at \SI{60}{\decibel} (measured at ear-level) using a MEG-compatible flat-panel speaker (Panphonics Oy, Espoo, Finland) positioned approximately \SI{2}{\meter} in front of the participant.

Additional physiological signals were recorded and synchronized with the MEG data, including electrooculography (EOG) for eye movements and electrocardiography (ECG) for heartbeat monitoring. Structural magnetic resonance images (MRI) were obtained separately on a Siemens 3T MRI scanner using a high-resolution T1-weighted MPRAGE sequence for precise anatomical reference.

\subsection{Audio Preprocessing and MEG-Audio Alignment}

Before training the machine learning models, several preprocessing steps were carried out to prepare the dataset thoroughly. All audio recordings were manually transcribed and verified multiple times to correct minor deviations from the original texts, such as pauses, repetitions, or speech errors. Additionally, each audio recording was manually cleaned to remove environmental and accidental participant-generated noises. Further noise reduction was conducted using Audacity 2.2.1 and Meta's Denoiser 0.1.5~\cite{defossez2020real}, ensuring the audio was as clean and artifact-free as possible.

Word and phone-level annotations for the cleaned audio were generated using the Montreal Forced Aligner~\cite{mfa}. Since the dataset is in Spanish, we used the Spanish Acoustic Model~\cite{mfa_spanish_mfa_acoustic_2022} along with an extended Spanish Dictionary model~\cite{mfa_spanish_spain_mfa_g2p_2022}. All the transcriptions, labels, and their alignments were manually inspected in Praat~\cite{Praat2009}.

Alignment between speech and MEG data was achieved using a dedicated \texttt{MISC} channel that simultaneously captured the microphone signals and reproduced audio alongside neural data. The audio in this channel, initially coarse and sampled at the MEG sampling frequency, allowed for signal synchronization. An alignment algorithm provided by the dataset authors was applied, comprising two main steps: an initial rough alignment via cross-correlation across a range of possible delays, followed by several refinement iterations. Each subsequent iteration improved synchronization by narrowing the delay range and progressively expanding the frequency band used for alignment, achieving precise MEG and audio data synchronization.

\subsection{Brain Signal Preprocessing and Filtering}\label{sec:meg_preproc}

Several signal processing techniques were employed to enhance the quality of neural data and improve model learning:

\begin{itemize}
    \item Only gradiometer sensors were used, excluding magnetometer data due to their higher susceptibility to environmental noise.
    \item Signals were decimated by a factor of 10, resulting in a sampling frequency of \SI{100}{\hertz}. A low-pass antialiasing filter with a cutoff frequency of \SI{50}{\hertz} was applied prior to decimation. This preprocessing step significantly reduces the dimensionality of the input data, making the model training computationally feasible and more efficient by decreasing memory usage and training time without sacrificing critical information contained within the signals.
    \item Band-pass filtering was performed using a FIR filter to retain frequencies between \SIrange{0.2}{31}{\hertz}, encompassing Delta (\SIrange{0.2}{3}{\hertz}), Theta (\SIrange{4}{7}{\hertz}), Alpha (\SIrange{8}{13}{\hertz}), and Beta (\SIrange{14}{31}{\hertz}) frequency bands. These frequency ranges correspond to various cognitive states relevant to speech processing~\cite{brain_bands}. Note: this filtering was only applied during model training for the model and task comparisons but not during the frequency band analysis (Section~\ref{sec:freqs}).
\end{itemize}

Wavelet-based denoising techniques were also applied due to their proven efficacy in reducing noise and specifically removing speech-related muscular artifacts from brain signals~\cite{Harender2017, Geetha2011, Dash2019}. Specifically, the discrete wavelet transform (DWT) was performed using Daubechies-4 wavelets with a two-level decomposition, as represented in Equation~\ref{equation:wavelet}.
Here, \(s\) denotes the original MEG signal. The wavelet decomposition splits \(s\) into approximation components (\(a_1\), \(a_2\)), which contain lower-frequency neural oscillations, and detail components (\(d_1\), \(d_2\)), capturing primarily higher-frequency artifacts and noise.

\begin{equation}
  s = a_1 + d_1 = a_2 + d_2 + d_1
  \label{equation:wavelet}
\end{equation}

The first-level decomposition produces the approximation \(a_1\) (frequencies below \SI{250}{\hertz}) and detail component \(d_1\) (\SIrange{250}{500}{\hertz}). A second decomposition further divides \(a_1\) into the lower-frequency approximation \(a_2\) (\SIrange{0.1}{125}{\hertz}) and detail component \(d_2\) (\SIrange{125}{250}{\hertz}). Since the higher-frequency components \(d_1\) and \(d_2\) predominantly contain muscular artifacts and noise, these were discarded. Consequently, only the approximation component \(a_2\), which captures the neural oscillations relevant for speech processing within the High-Gamma frequency band and below, was retained for subsequent analyses. Wavelet filtering was applied prior to decimation to effectively remove high-frequency noise and muscular artifacts from the original MEG signals sampled at \SI{1000}{\hertz}.

\subsection{Phone Distribution Analysis}

\begin{figure}[!tp]
  \centering
  \includegraphics[width=0.8\linewidth]{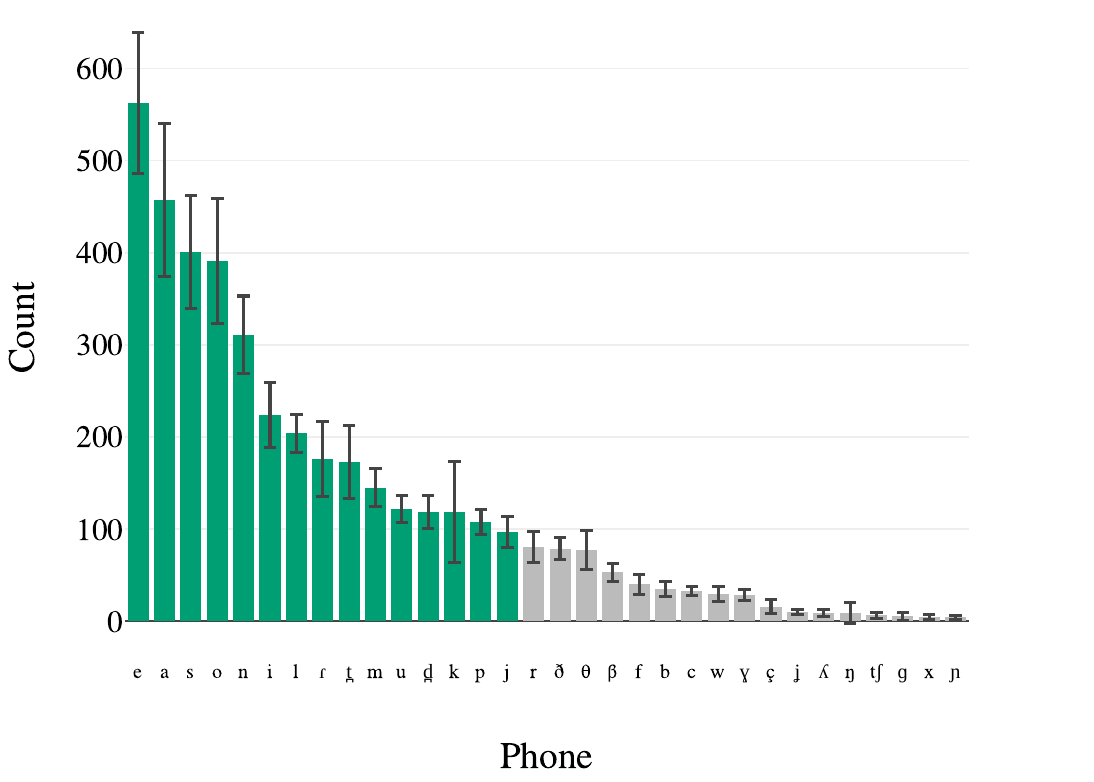}
  \caption{Average phone count in all subjects, including production, listening, and playback tasks. The black bars show their standard deviation. The phones selected for this work are marked in green, including five vowels and 10 consonants.}
  \label{fig:phone_counts}
\end{figure}

Following alignment, labels associated with segments not recorded by MEG, such as premature speech initiations or incomplete endings, were excluded. Figure \ref{fig:phone_counts} illustrates the average occurrence count of each phone in all recordings, considering the tasks performed by each participant (production, listening, and playback). Phones \textipa{[e]} and \textipa{[a]} were the most frequent, whereas \textipa{[x]} and \textipa{[\textltailn]} appeared least often. Note that this analysis refers to label distributions used to train the models.

Given the inherent phonetic imbalance of natural language speech, we extended our previous work from 10 phones to a broader selection of 15 phones. These include vowels and consonants frequently occurring in the recordings, ensuring at least 50 instances per phone class and a robust sample size for model training. This expanded selection provides better phonetic coverage and enriches the analysis by capturing more detailed phonetic distinctions.

\subsection{Epoching}

In neuroscience, epoching refers to extracting short segments of neural recordings around specific events, enabling analysis of how brain activity changes in response to particular stimuli or actions. Epochs were extracted around annotated phone events from the continuous MEG signals using the MNE library version 1.6.1~\cite{mne1,mne2}. Each epoch spans from \SI{100}{\milli\second} before to \SI{200}{\milli\second} after each phonetic event, encompassing the essential neural activity related to phonetic processing~\cite{phonetic_peak}. Baseline correction was applied using the pre-event segment to normalize the data and improve analytical reliability.

After epoch extraction, data were reshaped and balanced across phone pairs, ensuring equal representation of classes to mitigate class imbalance and allow fair comparative analyses.

\subsection{Classification Models}

The core objective of this work is to perform binary classification tasks, specifically targeting phonetic units (phones) in pairs. This pairwise classification strategy was adopted primarily due to the limited amount of data available per phone. By doing so, we aim to analyze which phones tend to be easily confused and which ones are more distinguishable based on brain recordings.

Given the dataset's size constraints, we selected a set of binary classification machine-learning models commonly used in cognitive neuroscience research~\cite{enc_dec_models}. These methods provide interpretability and computational efficiency, which are important characteristics when working with small datasets:

\begin{itemize}
    \item \textbf{Elastic Net}: A logistic regression model incorporating both L1 (Lasso) and L2 (Ridge) regularizations. Elastic Net combines the sparsity property of Lasso, which sets some feature weights to zero, with the regularization and stability properties of Ridge regression~\cite{elasticnet}. This approach is especially beneficial when dealing with correlated features, allowing multiple correlated variables to contribute to the model. We used the Scikit-learn Python library (version 1.3.2) for implementation~\cite{scikit-learn}.

    \item \textbf{Support Vector Machines (SVM)}: An effective classification method widely used for both linear and non-linear problems~\cite{Cortes1995, Vapnik1995}. Here, we employed the C-Support Vector Classification variant with a Radial Basis Function (RBF) kernel, also known as a Gaussian kernel~\cite{Scholkopf2002}. The RBF kernel allows SVMs to manage complex decision boundaries and is robust in high-dimensional feature spaces, thus making it particularly suitable for brain data classification tasks.

    \item \textbf{Linear Discriminant Analysis (LDA)}: A classical statistical method for classification that derives linear decision boundaries based on probabilistic assumptions. LDA calculates the likelihood that a data point belongs to each class and then applies Bayes' rule to assign class membership. It is computationally efficient, requires no parameter tuning, and provides intuitive interpretations of class separation~\cite{Fisher1936, Mclachlan2005}.
\end{itemize}

To include more recent modeling approaches, we extended our analysis by integrating neural network architectures recently applied in some neuroscience and speech-related tasks:

\begin{itemize}
    \item \textbf{Feed-forward Neural Networks (FFNs)}: We trained three variations of FFNs with differing depths~\cite{Rumelhart1986}:
    \begin{itemize}
        \item A single-layer FFN (FFN\(_{L=1}\)), consisting solely of a linear layer mapping inputs directly to output classes.
        \item A two-layer FFN (FFN\(_{L=2}\)) with one hidden layer of size 1024, utilizing rectified linear unit (ReLU) activation functions.
        \item A deeper three-layer FFN (FFN\(_{L=3}\)) with two hidden layers of 2048 and 1024 neurons, respectively. Deeper networks theoretically allow for capturing more complex feature interactions but may be prone to overfitting given limited data.
    \end{itemize}
    These FFNs serve as straightforward baseline models, helping us evaluate the trade-off between model complexity and available dataset size. By experimenting with various model depths and sizes, we aim to identify the most suitable architecture that balances performance and computational feasibility, especially given the constraints of our current dataset size.

    \item \textbf{Convolutional Neural Network (CNN)}: We implemented a simple CNN model specifically designed for dimensionality reduction and capturing local temporal patterns from the MEG signals~\cite{Lecun1998}. This model consists of a one-dimensional convolutional layer that applies convolution across each of the 204 sensor channels simultaneously. The convolutional layer uses a kernel size of 10 and a stride of 10, performing downsampling by a factor of 10 and reducing the temporal dimension from 31 time points to just 3 per sensor. The resulting feature representation, now compressed into a lower-dimensional embedding, is then directly fed into a linear classification layer that outputs predictions for the binary phone classification task. This approach is inspired by successful CNN-based architectures commonly used for processing continuous waveform signals in speech processing, such as WaveNet~\cite{Vandenoord2016} and various ASR models~\cite{Gulati2020, Radford2022}, which are able to capture temporal dependencies in audio data.

    \item \textbf{Transformer (DyslexNet-based model)}: We also implemented a Transformer encoder model inspired by DyslexNet~\cite{Klimovichgray2023}, a neural model initially designed to estimate semantic surprisal from brain signals using a Transformer architecture~\cite{Vaswani2017}. Specifically, we adapted DyslexNet by reimplementing it with a BERT-based encoder~\cite{Devlin2019} and applying two parameter-reduction techniques: cross-layer parameter sharing and factorized linear projections. Our Transformer consists of 4 shared encoder layers, each with a hidden dimension of 3072 units, an embedding dimension of 768 units (further factorized through an intermediate layer of 128 units), and 12 attention heads. Cross-layer parameter sharing considerably reduces the number of parameters, improving computational efficiency and limiting overfitting. The self-attention mechanism of this Transformer architecture can model long-range temporal dependencies in the MEG signal, possibly capturing more complex patterns of neural activation related to speech production.
\end{itemize}

All neural machine learning and neural models were trained under identical conditions for consistency and direct comparability across methodologies. This included identical preprocessing pipelines, namely decimation (downsampling), wavelet-based filtering, and Beta-band frequency filtering (\(\leq\) \SI{31}{\hertz}). Moreover, all neural network models were optimized using the Adam optimizer~\cite{Kingma2017} with a fixed learning rate of \(1 \cdot 10^{-4}\) and weight decay of \(1 \cdot 10^{-3}\), employing an early stopping criterion to prevent overfitting~\cite{Goodfellow2016, Dash2019}.

Before training the classification models, the input features underwent z-score normalization. This normalization involved subtracting the mean and dividing by the standard deviation for each feature, computed individually across all channels within each recording, subject, task, and phone pair. This step ensures that each feature contributes equally during the model training process, improving model stability and convergence.

\subsection{Frequency Band Analysis for Filter Selection}\label{sec:freqs}

To support the choice of frequency filtering in our main classification pipeline,  we performed a frequency band analysis to explore the relevance of distinct brainwave frequencies for speech decoding. Unlike the other experiments in this study, this analysis used only the gradiometer sensors without applying the other preprocessing steps, such as decimation, wavelet-based denoising, or low-pass filtering. This minimal preprocessing setup was chosen to avoid distorting the spectral properties of the original signal, enabling a fair comparison of decoding performance across distinct frequency bands.

We employed standard brainwave frequency bands commonly utilized in neuroscientific studies. These frequency bands provide a structured framework to investigate which portions of the neural signal carry critical information relevant for phonetic decoding~\cite{brain_bands, brain_hga}. Specifically, we analyzed six primary bands: Delta (\SIrange{0.2}{3} {\hertz}), Theta (\SIrange{4}{7} {\hertz}), Alpha (\SIrange{8}{13} {\hertz}), Beta (\SIrange{14}{31} {\hertz}), Gamma (\SIrange{32}{100} {\hertz}), and High-Gamma Activity (HGA, \SIrange{60}{300} {\hertz}). To isolate the frequency bands, we applied zero-phase finite impulse response (FIR) band-pass filters. These filters employed the default Hamming window with automatically adjusted transition bandwidths, ensuring effective suppression of frequencies outside the selected range while preserving the neural signal characteristics within each band. Filters were implemented separately for each band. This analysis allowed us to systematically identify the frequency ranges carrying the most decodable phonetic information for each speech modality. The results from this analysis were then used to motivate the application of the frequency low-pass filter in the model and task-comparison decoding experiments, which include full preprocessing and model training. The actual impact of this filter was then assessed through ablation testing.

\subsection{Ablation Study Design}

To evaluate the individual contributions of each preprocessing and filtering step used in our decoding pipeline detailed in Section \ref{sec:meg_preproc}, we conducted an ablation study. This analysis systematically measured the impact of removing specific components from the signal processing chain on the final decoding performance for speech production.

The baseline configuration included the full preprocessing pipeline: selection of gradiometer sensors, wavelet-based denoising, signal decimation (by a factor of 10), and a Beta-band low-pass filter (\(\leq\) \SI{31}{\hertz}). To measure each method's significance, we individually disabled each of the components from the baseline: sensor configuration (e.g., using magnetometers only), wavelet filtering, decimation, Beta-band filtering, and L1/L2 regularization terms.

The goal of this study was to isolate and quantify the influence of each individual step on the final decoding accuracy. This analysis allowed us to identify which preprocessing operations significantly and most strongly contribute to decoding performance and helped validate the final pipeline adopted for the rest of the experiments.

\subsection{Statistical Significance Analysis}

The statistical significance of the differences observed between the model evaluation metrics was assessed using the Wilcoxon signed-rank test~\cite{Wilcoxon1945}. This test was chosen because it is a non-parametric method well-suited for situations where the data distribution may deviate from normality, a scenario frequently encountered in studies with limited sample sizes. Specifically, the Wilcoxon signed-rank test evaluates whether the median difference between paired observations significantly differs from zero~\cite{Santafe2015}. For consistent and valid comparisons, we ensured identical samples across each model evaluation by employing a fixed 5-fold cross-validation split. To ensure valid paired comparisons, we used an identical fixed 5-fold cross-validation split across all models. Thus, each fold included exactly the same examples for every model, guaranteeing consistent pairing for the statistical analysis. A p-value threshold of less than 0.01 was used to determine statistical significance, ensuring robust evidence that observed differences are unlikely to have arisen purely by chance. Throughout this paper, whenever we refer to a \textit{difference}, we explicitly mean a statistically significant difference according to this criterion.

Accuracy was selected as the primary metric for statistical analysis due to its intuitive interpretability and its widespread use in classification studies. Nevertheless, for completeness and to facilitate comparison with other related studies, we also present the F-1 scores and the Area Under the Curve (AUC) alongside accuracy.

\section{Results}

\subsection{Model Performance}

Table \ref{tab:results_phone_pairs_self} presents the results obtained by each model for the speech production task, averaged across subjects, using 15-phone pairs. The plus/minus values indicate the standard deviation, representing the variability observed among different participants. The preprocessing steps employed for all models included selecting only gradiometer sensors (excluding magnetometer readings), applying a Beta band low-pass filter, performing wavelet filtering, and down-sampling the data by a factor of 10 as detailed in Section \ref{sec:meg_preproc}.

\begin{table}[!tbp]
  \caption{Model performance results for the speech production task (15-phone pairs).}
  \label{tab:results_phone_pairs_self}
  \centering
  \begin{tabular}{lrrr}
    \toprule
    \multicolumn{1}{c}{\textbf{Model}} & \multicolumn{1}{c}{\textbf{Accuracy (\%)}} & \multicolumn{1}{c}{\textbf{F-1 (\%)}} & \multicolumn{1}{c}{\textbf{AUC (\%)}} \\
    \midrule
    Elastic Net & \textbf{76.6 \(\pm\) 10.5} & \textbf{76.5 \(\pm\) 10.7} & \textbf{83.6 \(\pm\) 10.9} \\
    SVM & 75.1 \(\pm\) 10.6 & 74.8 \(\pm\) 11.0 & 82.2 \(\pm\) 11.1 \\
    LDA & 71.2 \(\pm\) 10.5 & 71.1 \(\pm\) 10.8 & 77.7 \(\pm\) 11.7 \\
    FFN\(_{L=1}\) & 74.9 \(\pm\) 10.8 & 74.6 \(\pm\) 10.9 & 81.5 \(\pm\) 11.6 \\
    FFN\(_{L=2}\) & 70.9 \(\pm\) 11.1 & 70.4 \(\pm\) 11.7 & 77.9 \(\pm\) 12.3 \\
    FFN\(_{L=3}\) & 65.6 \(\pm\) 11.7 & 64.3 \(\pm\) 13.0 & 73.3 \(\pm\) 13.5 \\
    CNN & 65.9 \(\pm\) 11.2 & 64.2 \(\pm\) 12.5 & 72.5 \(\pm\) 13.1 \\
    Transformer & 57.5 \(\pm\) \phantom{0}8.5 & 56.5 \(\pm\) \phantom{0}8.9 & 60.0 \(\pm\) 10.7 \\
    \bottomrule
  \end{tabular}
\end{table}

Overall, the Elastic Net model achieved the highest performance (accuracy: 76.6\%, F-1: 76.5\%, AUC: 83.6\%), significantly surpassing other classifiers, as confirmed by paired Wilcoxon signed-rank tests. Specifically, Elastic Net significantly outperformed both SVM (\(W=1.15\cdot10^7\), \(p<0.001\)) and LDA classifiers (\(W=3.81\cdot10^6\), \(p<0.001\)) in accuracy. Among neural models, the simplest single-layer feed-forward network (FFN\(_{L=1}\)) demonstrated performance comparable to SVM, yet still below Elastic Net (74.9\%, with \(W=1.15 \cdot 10^7\), \(p<0.001\)). Remarkably, increasing the complexity of the neural models, adding more layers (FFN\(_{L=2}\) and FFN\(_{L=3}\)) resulted in substantially decreased accuracy (70.9\% and 65.6\%, respectively, with \(W=1.96\cdot10^7\), \(p<0.001\) and \(W=1.33\cdot10^7\), \(p<0.001\)), likely due to overfitting given the limited data available. Similar trends were observed with convolutional (65.9\%, with \(W=7.26\cdot10^6\), \(p<0.001\)) and transformer-based (57.9\%, with \(W=1.10 \cdot 10^6\), \(p<0.001\)) architectures, further underscoring the challenge of training deep models with relatively small MEG datasets.

These results indicate that although neural network architectures can successfully learn to decode neural data related to speech production, their overall performance is lower than traditional regularized linear methods such as Elastic Net. This suggests that traditional machine learning approaches are still particularly effective in handling high-dimensional yet limited datasets typical in neuroimaging studies. Neural network models may require substantially more training data or additional regularization strategies (e.g., transfer learning~\cite{Kostas2020}, pretraining~\cite{Banville2021}, or data augmentation~\cite{Hartmann2018}) to achieve comparable or superior performance.

\subsection{Comparison Between Tasks}

\begin{table}[!bp]
  \caption{Results obtained using the Elastic Net model across speech modalities.}
  \label{tab:results_modalities_elastic}
  \centering
  \begin{tabular}{lrrr}
    \toprule
    \multicolumn{1}{c}{\textbf{Modality}} & \multicolumn{1}{c}{\textbf{Accuracy (\%)}} & \multicolumn{1}{c}{\textbf{F-1 (\%)}} & \multicolumn{1}{c}{\textbf{AUC (\%)}} \\
    \midrule
    Listening & 51.2 \(\pm\) \phantom{0}7.4 & 50.8 \(\pm\) \phantom{0}8.5 & 51.7 \(\pm\) \phantom{0}8.6 \\
    Playback & 51.2 \(\pm\) \phantom{0}7.1 & 51.0 \(\pm\) \phantom{0}8.0 & 51.7 \(\pm\) \phantom{0}8.2 \\
    Production & \textbf{76.6 \(\pm\) 10.5} & \textbf{76.5 \(\pm\) 10.7} & \textbf{83.6 \(\pm\) 10.9} \\
    \bottomrule
  \end{tabular}
\end{table}

Table \ref{tab:results_modalities_elastic} presents the results of the Elastic Net model applied across the three speech modalities, averaged by subjects and phone pairs. Each modality-specific model was trained separately for each subject and phone pair, applying identical preprocessing methods outlined earlier in Section \ref{sec:meg_preproc}. Particularly, the decoding performance in speech production is considerably higher compared to both speech perception modalities (listening and playback). These differences were confirmed to be statistically significant using the Wilcoxon signed-rank test, showing marked distinctions between production and listening (\(W=1.41\cdot10^5\), \(p<0.001\)) and between production and playback (\(W=8.66\cdot10^4\), \(p<0.001\)). Conversely, no significant difference was observed between listening and playback (\(W=2.19\cdot10^7\), \(p=0.523\)), consistent with the findings of the original dataset study~\cite{Bourguignon2020}.

Moreover, the decoding accuracies in all modalities significantly exceed the chance level (\(W=1.60\cdot10^7\), \(p<0.001\) for listening; \(W=1.60\cdot10^7\), \(p<0.001\) for playback; and \(W=1.64\cdot10^4\), \(p<0.001\) for production), indicating that the model indeed captures meaningful phonetic information in each modality.

\begin{figure}[!tp]
  \centering
  \begin{minipage}{0.49\textwidth}
      \centering
      \includegraphics[width=\linewidth]{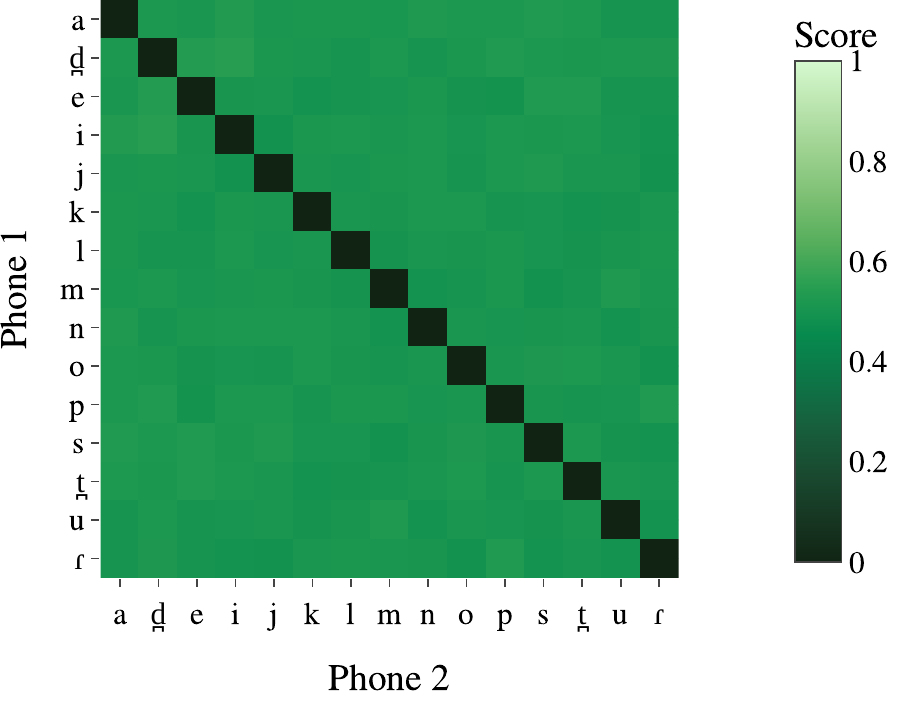}
      \caption{Averaged decoding accuracies for all the subjects on the listening task.}
      \label{fig:results_matrix_listen}
  \end{minipage}\hfill
  \begin{minipage}{0.49\textwidth}
      \centering
      \includegraphics[width=\linewidth]{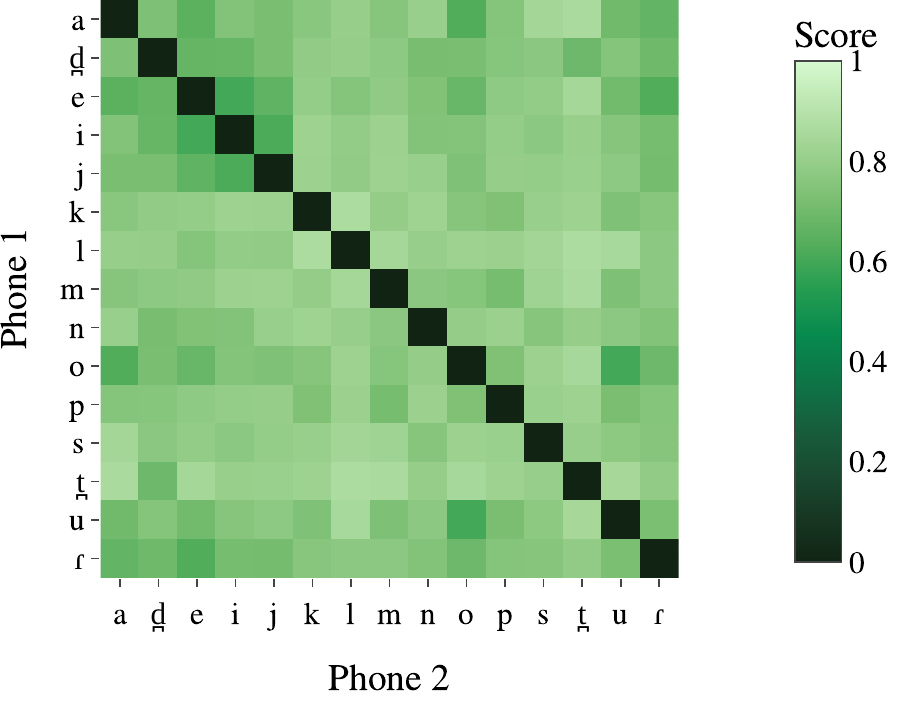}
      \caption{Averaged decoding accuracies for all the subjects on the production task.}
      \label{fig:results_matrix_self}
  \end{minipage}
\end{figure}

A more detailed comparison is visualized in Figures \ref{fig:results_matrix_listen} and \ref{fig:results_matrix_self}, showing the phone-pair classification accuracies across all subjects for listening and production, respectively. The matrices emphasize how production modality achieves higher accuracies, illustrated by lighter colors compared to listening. Specifically, the highest decoding accuracy for speech production was achieved for the phone pair \textipa{[l]}-\textipa{[\textsubbridge{t}]} (87.08\%), and even the lowest accuracy, observed for the vowel pair \textipa{[e]}-\textipa{[i]} (60.16\%), exceeds the best accuracy obtained during listening. In the listening modality, the highest and lowest accuracies were observed for the pairs \textipa{[\textsubbridge{d}]}-\textipa{[i]} (54.65\%) and \textipa{[i]}-\textipa{[j]} (48.86\%), respectively, underscoring the modality differences in neural decoding abilities.

\subsection{Frequency Band Contribution to Phonetic Decoding}

\begin{figure}[!t]
    \centering
    \includegraphics[width=0.8\linewidth]{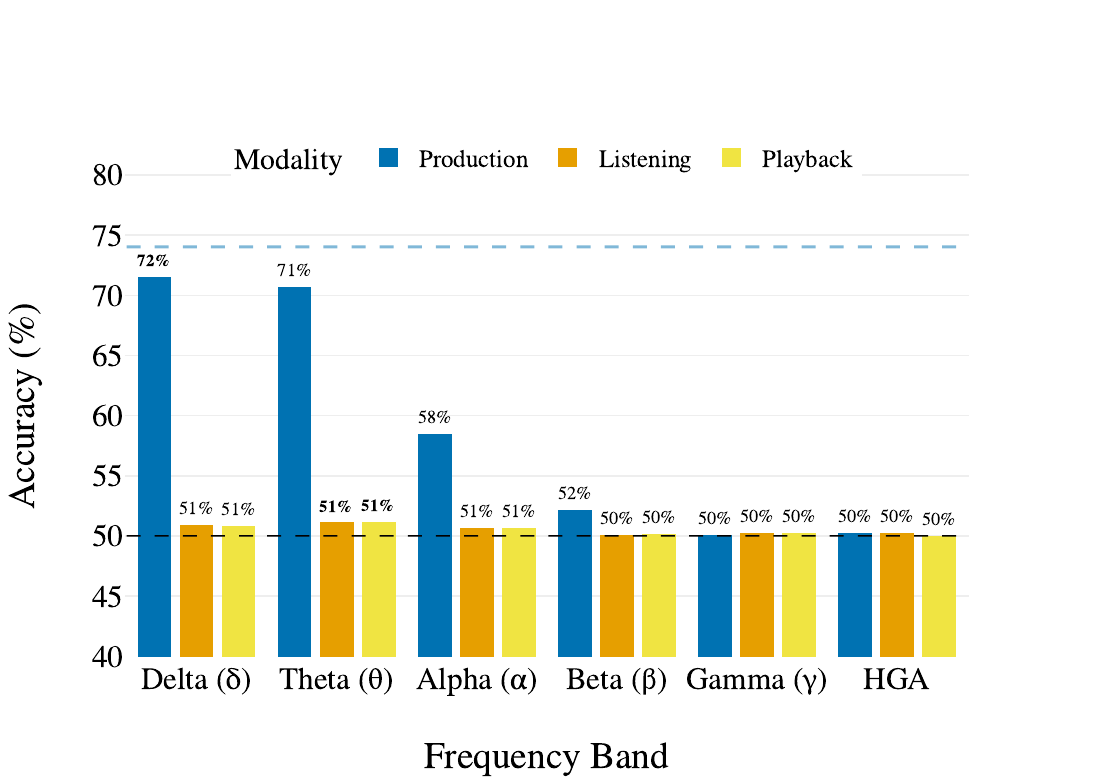}
    \caption{Decoding accuracies by frequency band for speech perception (Listening and Playback) and speech production modalities. The plot y-axis ranges from 40\% to 80\% for better visibility. Bold annotations indicate the best accuracy for each modality. The dashed black line represents the chance accuracy level (50\%), and the dashed blue line indicates the baseline (ceiling) accuracy for the production modality (74.02\%), obtained without frequency filtering.} \label{fig:freq_results_modailities}
\end{figure}

Figure \ref{fig:freq_results_modailities} summarizes the decoding accuracies across different frequency bands for speech perception (listening and playback) and speech production. These models were trained using Elastic Net regression and with minimal preprocessing: without decimation (and thus without antialiasing filters), without low-pass filters, and without wavelets filtering to avoid inadvertently affecting frequency band characteristics.

In the speech production modality, decoding accuracies were notably higher for low-frequency oscillations (Beta range and lower), demonstrating their relevance for phonetic decoding during articulation. The highest decoding accuracy for speech production was achieved using the Delta band (71.56\%), closely followed by Theta (70.72\%), both statistically significant compared to random chance (\(W=5.11 \cdot 10^4\), \(p<0.001\) and \(W=9.38 \cdot 10^4\), \(p<0.001\), respectively). Delta-band oscillations alone accounted for 96.7\% of the maximum accuracy obtained using all frequencies (baseline accuracy: 74.02\%, indicated by the dashed blue line). These results emphasize the critical role of slow cortical oscillators in speech production.

In contrast, decoding accuracies for speech perception modalities (listening and playback) were generally lower but remained above chance level in the Alpha and lower frequency bands. The best decoding performance for listening was observed with the Theta band (51.16\%, \(W=1.60 \cdot 10^7\), \(p<0.001\)), representing approximately 99.96\% of the baseline (all frequencies) accuracy for listening (51.18\%). Similarly, playback reached its maximum accuracy with the Theta band (51.20\%, \(W=1.60 \cdot 10^7\), \(p<0.001\)), accounting for 100\% of the playback baseline. These percentages, which indicate near-baseline performance, were not statistically significant compared to the baseline results (\(W=1.85 \cdot 10^7\), \(p=0.970\) for listening, and \(W=1.86 \cdot 10^7\), \(p=0.524\) for playback), which were obtained without frequency filtering. In particular, decoding from frequencies higher than Alpha (Beta, Gamma, HGA) yielded very limited performance, with accuracies approaching random chance (50\%), suggesting either minimal relevant phonetic information or possible limitations of our linear decoding model.

Overall, there is substantial overlap in relevant frequency ranges between speech perception and production, particularly within low-frequency oscillations (Delta, Theta, and Alpha bands). However, speech production decoding clearly benefits from information present in frequencies extending up to the Beta band. The reduced accuracy at higher frequencies, such as the Gamma and High-Gamma activity (HGA) ranges, could indicate either the absence of relevant speech information in these bands or the inability of linear models to capture higher-frequency oscillations or more complex brain activity patterns.

Based on these results, we opted to apply a Beta-band low-pass filter (\(\leq\) \SI{31}{\hertz}) in all model training experiments. This decision was motivated by the fact that, in speech production, phonetic information was decodable from Delta through Beta frequencies, while in speech perception, informative content was limited to Delta through Alpha bands. The Beta-range filter, therefore, preserves all relevant frequency components across modalities. The actual contribution of this filter to decoding performance is quantitatively evaluated in the following ablation study.

\subsection{Ablation Study}

An ablation study was conducted to assess the impact of each preprocessing and sensor configuration on the decoding performance for speech production. The baseline model used Elastic Net regression, gradiometer sensors, wavelets filtering, decimation (factor of 10), and a Beta-band low-pass filter (\SI{\leq 31}{\hertz}). The results of this study are presented in Table \ref{tab:results_ablation} and are summarized below.

\begin{table}[!tp]
  \caption{Ablation results on speech production modality. Bold values indicate the best performance.}
  \label{tab:results_ablation}
  \centering
  \resizebox{\columnwidth}{!}{
  \begin{tabular}{lrrr}
    \toprule
    \multicolumn{1}{c}{\textbf{Configuration}} & \multicolumn{1}{c}{\textbf{Accuracy (\%)}} & \multicolumn{1}{c}{\textbf{F-1 (\%)}} & \multicolumn{1}{c}{\textbf{AUC (\%)}} \\
    \midrule
    \textit{Full Model (baseline)} & \textbf{76.6 \(\pm\) 10.5} & \textbf{76.4 \(\pm\) 10.7} & \textbf{83.5 \(\pm\) 10.9} \\
    \midrule
    Magn. sensors only  & 72.0 \(\pm\) 10.6 & 71.8 \(\pm\) 10.9 & 78.5 \(\pm\) 11.7 \\
    Magn.+Grad. sensors  & 75.9 \(\pm\) 10.5 & 75.7 \(\pm\) 10.8 & 82.8 \(\pm\) 11.0 \\
    No wavelets filter  & 75.4 \(\pm\) 10.8 & 75.3 \(\pm\) 11.0 & 82.3 \(\pm\) 11.3 \\
    No decimation  & 75.2 \(\pm\) 10.8 & 75.0 \(\pm\) 11.1 & 82.0 \(\pm\) 11.4 \\
    No L1 (Ridge)  & 70.6 \(\pm\) 10.1 & 70.4 \(\pm\) 10.4 & 77.1 \(\pm\) 11.3 \\
    No L2 (Lasso)  & 75.4 \(\pm\) 10.5 & 75.3 \(\pm\) 10.8 & 82.3 \(\pm\) 11.1 \\
    No Beta filter (\(\leq\)31 Hz) & \textbf{76.6 \(\pm\) 10.5} & \textbf{76.4 \(\pm\) 10.7} & \textbf{83.5 \(\pm\) 10.9} \\
    \bottomrule
  \end{tabular}
  }
\end{table}

Using magnetometer sensors alone (excluding gradiometers) significantly reduced accuracy, dropping to 72.0\ (\(W=3.41\cdot10^6\), \(p<0.001\)), marking a substantial performance loss of approximately 6\%. Including both magnetometers and gradiometers slightly mitigated this drop, achieving an accuracy of 75.9\%, though this still represented a significant reduction from the baseline (performance loss: 1\%, \(W=7.58\cdot10^6\), \(p<0.001\)). While gradiometer and magnetometer signals are both reconstructed from a common signal space under Maxwell filtering, these results suggest that, in our decoding framework, classifiers perform better when trained exclusively on gradiometer data. This likely reflects the practical difficulties faced by our classifiers in extracting information from the different sensor types, rather than the inherent differences in information content, and underscores the importance of sensor configuration in the performance of the BCI model.

Removing wavelet filtering also had a considerable negative impact, reducing decoding accuracy to 75.4\% (performance loss: 2\%; \(W=1.12\cdot10^7\), \(p<0.001\)). Although wavelet decomposition focuses on discarding detail coefficients above \SI{125}{\hertz}, its time-frequency localization can also attenuate transient artifacts within lower frequency ranges that are not fully captured by standard FIR filtering alone. Hence, omitting wavelet-based denoising eliminates this additional layer of artifact suppression, thereby reducing classification performance. Likewise, removing decimation decreased accuracy significantly to 75.2\% (performance loss: 2\%; \(W=1.08\cdot10^7\), \(p<0.001\)).

Eliminating the L1 penalty (thereby retaining only L2, i.e., Ridge regression) led to the most significant performance drop, reducing accuracy to 70.6\% (an 8\% loss, \(W=3.14\cdot10^6\), \(p<0.001\)). This finding suggests that the sparse feature selection offered by L1 regularization is particularly advantageous in high-dimensional MEG data. Meanwhile, eliminating the L2 penalty (thus using a pure Lasso model) reduced accuracy to 75.4\% (2\% loss, \(W=6.37\cdot10^6\), \(p<0.001\)), indicating that while L2 stabilizes the weight distribution, it appears less critical than L1 in capturing the most relevant signals for speech production. Overall, these observations support the synergy of L1 and L2 regularizations when decoding phonetic units from MEG recordings, as realized by the Elastic Net baseline.

In contrast, removing the Beta-band low-pass filter did not significantly affect the decoding accuracy, maintaining performance at 76.6\%, identical to the baseline (\(W=7.19\cdot10^5\), \(p=0.184\)). A plausible explanation is that the decimation process itself inherently applies a low-pass filter (approximately \SI{50}{\hertz}), overlapping with the Beta range (\SI{\leq 31}{\hertz}), and that little or no crucial speech-related content resides in \SIrange{31}{50}{\hertz}. Thus, any additional filter restricting frequencies below \SI{50}{\hertz} may not significantly remove information beyond what the decimation filter already eliminates.

Overall, these findings highlight the essential contributions of gradiometer sensors, wavelet filtering, decimation (and its associated low-pass filtering), and L1/L2 regularizations to achieving optimal decoding performance. Among these, L1 regularization and sensor configuration (magnetometer exclusion) appear most critical, each independently contributing significantly to capturing essential neural information for decoding speech production processes.

\section{Conclusions}

This study demonstrates significant differences in decoding performance between speech production and speech perception tasks, with speech production yielding considerably higher decoding accuracies. Our best-achieved decoding accuracy of 76.6\% during speech production markedly contrasts with the near-chance accuracies ($\sim$51\%) obtained in passive listening and playback tasks. These results align with and extend our previous findings, confirming that neural signals during overt speech contain robust phonetic information and are decodable.

Expanding the analysis from 10 to 15 phonetic pairs increased the coverage of phonetic variability, thereby reinforcing the robustness of our decoding models across a broader range of phonetic contrasts. The superior decoding performance observed in speech production tasks likely arises from richer neural activation patterns related to motor planning, articulatory execution, and proprioceptive feedback, which are inherently stronger and more consistent during overt speech compared to passive auditory tasks.

The introduction of neural network models, including feed-forward, convolutional, and Transformer-based architectures, provided new insights into the potential of advanced computational methods for decoding MEG signals. Although these neural models are able to learn phonetic distinctions, their decoding accuracies consistently fall short of the regularized linear Elastic Net model. Particularly, increasing model complexity, especially in deeper feed-forward networks, convolutional neural networks, and transformer architectures, led to a substantial performance decrease, likely due to overfitting, underscoring the challenges posed by limited neuroimaging datasets. These findings emphasize that traditional machine-learning approaches, when combined with effective regularization, remain advantageous when dealing with small, high-dimensional datasets typical in cognitive neuroscience.

Furthermore, our analysis of brain signal frequency bands revealed distinct neural oscillatory contributions to speech decoding. Low-frequency bands (Delta and Theta) yielded the highest decoding performance during speech production, accounting for nearly all the accuracy achievable with broader frequency ranges. Conversely, higher-frequency bands (Gamma and High-Gamma) produced accuracies near chance level, suggesting either a limited presence of phonetic information or the inability of linear decoding models to capture high-frequency neural dynamics. These results point to the need for future research exploring non-linear or deep-learning models specifically adapted to detect more subtle neural patterns in higher-frequency activity.

Lastly, despite applying established wavelet-based filtering techniques to mitigate muscular artifacts, it is still uncertain whether our models decode exclusively neural activity or whether residual electromyographic artifacts inadvertently contribute phonetic information. In particular, it is unclear whether movement artifacts contributed to the decoding accuracy. Further methodological advancements are needed to more effectively disentangle genuine neural signals from possible interference.

Overall, this study emphasizes the importance of focusing on overt speech production paradigms for developing speech-based BCIs, highlighting traditional linear models as particularly effective, given current dataset limitations. Future research should focus on improving model generalizability, incorporating larger and richer datasets, and refining signal processing techniques to support the development of effective communication technologies for individuals with severe speech impairments.

\section*{Acknowledgements}

The authors sincerely thank the DIPC Supercomputing Center for providing computational resources and critical technical assistance throughout this research. We also gratefully acknowledge support from the IKUR-IKA-23/18 project. Finally, we thank Ekain Arrieta and Ander Barrena for generously sharing the implementation of the DyslexNet model with us.

\section*{Funding}

This work was supported by the IKUR-IKA-23/18 project.

\section*{Declaration of Generative AI and AI-Assisted Technologies in the Writing Process}

During the preparation of this work the author(s) used DeepSeek and ChatGPT (GPT-4, OpenAI) in order to correct grammar and improve the fluency of some sentences. After using these services, the authors reviewed and edited the content as needed and take full responsibility for the content of the published article.

\bibliographystyle{spmpsci}
\bibliography{references}

\end{document}